\documentclass{IEEEtran}
\usepackage{cite}
\usepackage{amsmath,amssymb,amsfonts}
\usepackage{algorithmic}
\usepackage[ruled,vlined]{algorithm2e}
\usepackage[caption=false]{subfig}
\usepackage{multirow}
\usepackage{graphicx}
\usepackage{textcomp}

\def\BibTeX{{\rm B\kern-.05em{\sc i\kern-.025em b}\kern-.08em
    T\kern-.1667em\lower.7ex\hbox{E}\kern-.125emX}}
\begin{document}

\newcommand{\JK}[1]{\textcolor{blue}{\textbf{JK: {#1}}}}
\newcommand{\John}[1]{\textcolor{blue}{\textbf{John: {#1}}}}

\title{Improving Botnet Detection with Recurrent Neural Network and Transfer Learning}

\author{\IEEEauthorblockN{
Jeeyung Kim\IEEEauthorrefmark{1},
Alex Sim\IEEEauthorrefmark{1},
Jinoh Kim\IEEEauthorrefmark{2},
Kesheng Wu\IEEEauthorrefmark{1},
Jaegyoon Hahm\IEEEauthorrefmark{3} \\
}
\IEEEauthorblockA{\IEEEauthorrefmark{1}Lawrence Berkeley National Laboratory, Berkeley, CA, USA,
\{jeeyungkim,asim,kwu\}@lbl.gov}\\
\IEEEauthorblockA{\IEEEauthorrefmark{2}Texas A\&M University-Commerce, Commerce, TX, USA,
jinoh.kim@tamuc.edu}\\
\IEEEauthorblockA{\IEEEauthorrefmark{3}KISTI, Daejeon, South Korea, 
jaehahm@kisti.re.kr}
}

\maketitle


\begin{abstract}
Botnet detection is a critical step in stopping the spread of botnets
and preventing malicious activities.
However, reliable detection is still a challenging task, due to a wide variety of botnets involving ever-increasing types
of devices and attack vectors.
Recent approaches employing machine learning (ML) showed improved
performance than earlier ones, but these ML-based approaches
still have significant limitations.
For example, most ML approaches can not incorporate
sequential pattern analysis techniques key to detect some classes
of botnets.
Another common shortcoming of ML-based approaches is the need to retrain
neural networks in order to detect the evolving botnets;
however, the training process is time-consuming and requires significant efforts to label the training data.
For fast-evolving botnets, it might take too long to create sufficient training samples before the botnets have changed again.
%
To address these challenges, we propose a novel botnet detection method,
built upon Recurrent Variational Autoencoder (RVAE) that effectively
captures \emph{sequential characteristics} of botnet activities. 
In the experiment, this semi-supervised learning method achieves better detection accuracy than similar learning methods, especially on hard to detect classes.
Additionally, we devise a \emph{transfer learning framework} to learn from a
well-curated source data set and transfer the knowledge to a target
problem domain not seen before. 
Tests show that the true-positive rate (TPR) with transfer learning is higher than the RVAE semi-supervised learning method trained using the target data set (91.8\% vs. 68.3\%). 
\end{abstract}



\section{Introduction}
\label{sec:introduction}
Botnet is one of the most significant threats to the cyber-security as they are the source of many malicious activities~\cite{alieyan2017survey}.
The compromised machines in a botnet are typically hijacked without the owner's knowledge.
These machines (also known as ``bots'') are then commanded to act together to attack targeted servers or find more valuable zombie candidates.
In fact, the botnet is frequently used to perform a variety of attacks, including distributed denial-of-service attacks (DDoS), click-fraud, spamming and crypto-mining.
Botnets also  harbor malware and ransomware for delivery to
additional victims.
A critical task of cybersecurity  research is to detect botnets and stop their attacks.

It is challenging to identify botnets due to following reasons.
First, the malicious software that infects a victim host and operates the botnet is evolving to evade detection.
For example, botnets may use a combination of communication protocols, such as Internet Relay Chat (IRC), peer-to-peer (P2P), and HTTP, rather than relying on one protocol~\cite{zhao2013botnet}.
Second, newly introduced botnets are capable of utilizing diverse types of computing devices and attack vectors. 
For example, in 2016 the Mirai botnet controlled hundreds of thousands
of Internet of Things (IoT) devices to conduct a high-profile DDoS
attack~\cite{antonakakis2017understanding}, and  in 2018
 another sophisticated botnet system called Smominru started crypto-mining.
These fast-evolving characteristics significantly reduce the effectiveness of the traditional detection approaches.

The existing botnet detection approaches fall into two broad categories: honeypots and Intrusion Detection Systems (IDS).
Honeypot is a computer system used as a trap to draw the attention of attackers~\cite{zeidanloo2010taxonomy}.
This approach has several limitations in scale and  detection capability~\cite{zeidanloo2010taxonomy}.
IDS monitors network systems to uncover malicious activities based on either pattern matching using predefined rulesets (``signature-based'') or behavioral analysis to discriminate anomalies from normal actions (``anomaly-based''). 
The signature-based method is configured with a set of rules or signatures to classify types of network traffics.
This approach often requires a relatively small amount of computation (e.g., using a customized pattern machine engine for faster look-up),  and
work in real-time without slowing down normal network operations.
The effectiveness of the signature-based detection has widely been studied, but it is only able to identify well-known botnets~\cite{roesch1999snort, zeidanloo2010taxonomy}.

An anomaly-based approach identifies network traffic anomalies, such as high network latency, high volumes of traffic, and unusual system behavior, to detect malicious actions~\cite{zeidanloo2010taxonomy}.
A common strategy is to model normal behavior of network traffic and then  declare significant statistical deviations as potential threats~\cite{abou2020botchase}.
Many statistical techniques and heuristics have been proposed to characterize
botnet anomalies~\cite{binkley2006algorithm, gu2008botsniffer}.
Recently, a number of machine learning (ML) techniques have shown to be very effective, even possible to detect previously \emph{unseen}
types of botnet attacks~\cite{venkatesh2012http, singh2014big, beigi2014towards, stevanovic2014efficient, zhao2013botnet}.

While attractive, there are several limitations with the existing ML-based
botnet detection algorithms.
One of the critical limitations is that these algorithms 
did not pay much attention to \emph{sequential patterns} 
within network data, even though botnet traffic shows
repeated patterns due to the nature of the pre-programmed
activities~\cite{assadhan2009detecting}.
There are studies considered sequential characteristics by the same source IP addresses, which may not be generalized to other IP addresses~\cite{sinha2019tracking, torres2016analysis}.
In addition, existing studies only consider specific types of network
activities, such as IRC, P2P, and HTTP traffic, while active botnets are
utilizing a combination of different protocols~\cite{nguyen2019gee, nicolau2018learning, sinha2019tracking,torres2016analysis}.

Another limitation of the existing approaches is the need of labeled training datasets.
The learning process requiring labeled training data is commonly known as supervised learning.
So far, anomaly detection based on supervised learning has achieved a high accuracy for detecting botnets~\cite{singh2014big, ongun2019designing, du2019fenet}.
The key drawback of supervised learning is the labeled training data is often unavailable in practice, especially, the malicious samples.
The \emph{semi-supervised learning} approach addresses this difficulty by requiring only one type of samples, typically, normal samples, for training.  
Popular semi-supervised learning algorithms include autoencoders
(AEs)~\cite{dargenio2018exploring}, Variational Autoencoder
(VAEs)~\cite{nguyen2019gee, nicolau2018learning,an2015variational},
one-class support vector machines (OSVMs)~\cite{nicolau2018learning}, and so on.

Another possible solution to address the shortage of the labeled data is \emph{transfer learning}, which utilizes labeled data available in one domain (``source domain'') for the domain of interest (``target domain'').
When there is insufficient labeled data in the target domain, transfer learning allows us to construct a learning model without the expensive data-labeling effort~\cite{pan2009survey}. 
So far, transfer learning has been successfully applied to
text classification, speech recognition, and image
classification~\cite{andrews2016transfer,
  chalapathy2018anomaly, ide2017multi, xiao2015robust}.
There are also examples of applying transfer learning for botnet
detection~\cite{Alothman:2018:SBI, Alothman:2018:CBS}. 
However, these techniques require high computational costs and only achieve lower performance compared to supervised and semi-supervised techniques.

This work has two main goals. 
First, we propose an ML method capable to represent sequential patterns within network data, by utilizing Recurrent Variational Autoencoder (RVAE).
From published literature, we know that sequential patterns of activities are critical for detecting many botnets~\cite{sinha2019tracking, torres2016analysis}.
The proposed method also includes a new {\em anomaly scoring} function that allows on-line detection of botnets.
Second, we introduce a new training framework based on transfer learning to facilitate the deployment of our botnet detection method without first collecting and labeling all possible anomalies in the target domain. 

The main contributions of this paper can be summarized as follows:
\begin{itemize}
    \item We present a new ML model for botnet detection built on Recurrent Variational Autoencoder (RVAE) to capture sequential patterns.  This is a semi-supervised approach that learns from the normal data and detects potential anomalies that may vary over time.
    \item We devise a scoring strategy that allows us to identify anomalies in real-time.  This scoring approach utilizes the probability density function of reconstruction errors from the RVAE network.
    \item We verify that our approach could detect changing
    botnets by splitting the popular test data set CTU-13 into training
    and testing sets with different types of botnets.  Tests show that we
    are able to detect botnets effectively when
    previously unseen types of botnets appear in testing but not in training.
    \item We propose a transfer learning framework to construct a learning model without the labeled training samples from the target domain.
    \item We verify that our transfer learning approach could detect potential botnets in a new network monitoring data set (as the target domain) with the knowledge transferred from the popular data set CTU-13 (as the source domain).
    The experimental results show that the presented method detects suspicious botnet connections effectively.
\end{itemize}

The remainder of paper is composed of 5 sections.  In Section 2, we review previous studies related to botnet detection and transfer learning.
In Section 3, we present our botnet detection methods, including both RVAE and transfer learning framework.
The overall botnet detection process consists of three steps: data pre-processing, anomaly scoring, and anomaly detection.
The transfer learning mechanism could work with or without labeled samples from the target domain.
We describe data sets and experimental setting in Section 4, and report our experimental results of anomaly detection and transfer learning in Section 5.
Finally, we give concluding remarks and possible future research directions of this work in Section 6. 

\section{Related work} 
Following a broad introduction in the previous section, we next provide more details on a number of related works that lead to the unique combinations of techniques to be described in Section 3.

\subsection{ML Methodology}
\noindent\textbf{Variational Autoencoder (VAE).}
Given an input $x$, the autoencoder neural network structure consists of a encoder network and a decoder network used together, where the encoder extracts the latent variable $z$ and the decoder reconstructure the input $x$ from $z$.
VAE use using a reparameterization trick to allow this autoencoder structure to follow the variational Bayesian inferencing theory~\cite{kingma2013auto}. 

Throughout this paper, we use $\theta$ to denote parameters of the encoder of VAE and $F$ to represent the encoder, $\phi$ to denote parameters of the decoder of VAE and $G$ to represent the decoder.
Let $x_n$ denote the $n$th input data record, we use $\tilde{x}_{n}$ to denote the output produced by the decoder $G$.
The loss function as follows:
\begin{equation}\label{eq:1}
    J(x) = - \mathbb{E}_{q_\phi(z|x)}[log p_\theta(x|z)] + \beta * D_{KL}[q_\phi(z|x)|p_\theta(z)] 
\end{equation}
$p$ represents the encoder, and $q$ represents the decoder.

\noindent\textbf{Gated Recurrent Unit (GRU).}
GRU is a Recurrent Neural Network (RNN) structure containing directed cycles for representing sequential patterns.  GRU is capable of remember long sequences,
which allows it to extract features across time~\cite{chung2014empirical}.

\noindent\textbf{Recurrent Variational Autoencoder (RVAE).}
RVAE is a structure that combines sequence-to-sequence model (seq2seq) with a VAE. The encoder and decoder of VAE consists of an auto-regressive model
As it utilizes RNN instead of multilayer perceptron (MLP) as the decoder, it takes the current input as well as its neighborhood into account to generate sequential outputs. 
The more detailed discussion of Recurrent VAE is available in literature~\cite{bowman2015generating, roberts2018hierarchical}.
Based on published literature on extracting patterns from text and music, RVAE is very effective for extract sequential patterns.
We expect this neural network structure could effectively extract sequential patterns critical for detect anomalies in network traffic as well.

\noindent\textbf{Transfer Learning.}
Transfer learning trains a model in one problem domain, named the source domain, with plenty of training data, and then apply the model to another application domain, named the target domain, where there is limited or insufficient amount of train data~\cite{pan2009survey}.  This model could be either classification or regression. 
The transfer learning can be divided into three categories according to the label existence in the source/target domains  and the types of tasks.

\emph{Inductive transfer learning} involves the same source and target domain, but different tasks.
\emph{Transductive transfer learning} involves different source and target domains, but, the same task in source and target domain.  Typically, the source domain and the target domain are closely related.
\emph{Unsupervised transfer learning} involves different source and target domains as well as different tasks.
The technique to be described later is a transductive transfer learning techniques.

\subsection{Botnet Detection Methodology using ML}
Many methods have been developed for botnet detection.
Here, we briefly describe a number of commonly used signature-based methods and ML techniques(VAE, RNN) to provide a more detailed context for our work.

\noindent\textbf{Signature-based Network Intrusion Detection System.}
Signature-based network IDS methods has been widely studied~\cite{zeidanloo2010taxonomy, roesch1999snort,abou2020botchase,binkley2006algorithm,gu2008botsniffer,paxson1999bro}.
In~\cite{abou2020botchase}, the authors design a botnet decision engine that determines any divergence or statistical deviations, which are based on normal network behaviors, over network traffic data.
Among this class of methods, Zeek is one of the most popular Network Intrusion Detection System (IDS), which is a monitoring system for detecting network intruders in real-time by passively monitoring a network link over which the intruder's traffic transits~\cite{paxson1999bro}.
Zeek analyzes packet traces (PCAP)  by utilizing {\tt libpcap}, the packet-capture library.
The system is divided into {\em event engine}, which reduces a stream of packets to a stream of higher-level network events, and {\em Policy Script Interpreter}, which logs real-time notifications and records data to disk.
Even though Zeek is not built for detecting botnet, we found that certain features it detects is highly correlated with many classes of botnets and plan to use these features to measure the detection performance of transfer learning method when we have no labels on the data records.

\noindent\textbf{Botnet Detection Using Variational Autoencoder.}
\cite{nguyen2019gee} introduced VAE as an unsupervised method for detecting anomalies and proposed to explain anomalies with a gradient-based fingerprinting technique.  However, they assumed that they already knew the ratio of anomaly and it did not consider sequential patterns that can increase the performance of detection.
Nicolau et al.~\cite{nicolau2018learning} proposed a revised VAE structure, called as Dirac Delta VAE, for better anomaly detection performance.
It narrowed down the range of latent space that makes classifiers detect anomaly easier.
However, Dirac Delta VAE cannot be trained end-to-end because it separates the classifier and the feature extractor. Furthermore, the authors conducted experiments using only one type of botnets for training and testing. 
There are various of studies utilizing VAE for anomaly detection of network traffic~\cite{an2015variational, kumagai2019transfer}, most of them overlooked sequential characteristics within network traffic.

\noindent\textbf{Botnet Detection Using Recurrent Neural Network.}
RNN is well-known for capturing sequential characteristics of network traffic data. 
For example, Sinha et al.~\cite{sinha2019tracking} proposed a supervised approach to detect botnet hosts by tracking the network activities over time and extract graph-based features from NetFlow data for botnet detection. 
However, their technique only extracted features for each host IP address separately. 
In addition, the method is trained and tested on the restricted botnet scenarios.
Torres et al.~\cite{torres2016analysis} assigned symbols to features such as port number and packet size, and embed these symbols in a distributed representation similar to word embedding.
Their work demonstrated the potential of using RNN for botnet detection, but their tests showed poor performance when the classes of different traffics are highly imbalanced. 
Ongun et al.~\cite{ongun2019designing} utilized text output from IDS as input to RNN.  However, their RNN was specific designed for text data. 
While many of these methods showed improved detection performance compared earlier generations of tools.  However, most of them are not able to perform on-line anomaly detection because they need to collect every traffic related to the host IP addresses in order to distinguish whether they are malicious or not.
In addition, existing ML-based methods usually require fully labeled data set that are hard to obtain due to lack of labeled data on changing network traffic.


\noindent\textbf{Botnet Detection Using Other Machine Learning Approach.}
Besides VAE and RNN, many recent studies have attempted to make use of various ML approach to reduce the dependence for human heuristics.
Du et al. regard every feature as sentence and embed the features. With embedded features, classifier can be trained to detect malicious botnets\cite{du2019fenet}. 
Singh et al. introduce the framework for P2P botnet detection using Random Forest (RF)\cite{singh2014big}.
Furthermore, Ongun et al. present how to extract features that are good for ML model. The authors use a statistics aggregated feature processing method and validate the method with RF and Gradient Boosting\cite{ongun2019designing}.

\noindent\textbf{Botnet Detection Using Transfer Learning.}
Some studies suggested making use of transfer learning on botnet detection~\cite{kumagai2019transfer, Alothman:2018:SBI, taheri2018leveraging, bhodia2019transfer, singla2019overcoming, jiang2019new}.
However, the method that is suggested by Alothman et al. depends on naive techniques such as calculating similarity between each instance in the source and the target domain, which requires high computation cost\cite{Alothman:2018:SBI}.
Moreover, Jiang et al. use clustering technique and naive rule methods and only focus on the botnet using Command and Control channel\cite{jiang2019new}. 
These methods are limited in that they cannot provide end-to-end learning manner while the proposed method can.
Furthermore, contrary to transfer learning studies in other area that use Deep Neural Networks structure~\cite{andrews2016transfer, chalapathy2018anomaly,ide2017multi}, the studies on botnet detection utilize relatively simple methods.

A few methods use Neural Networks in transfer learning~\cite{bhodia2019transfer, taheri2018leveraging,singla2019overcoming}. 
The authors treat network traffic features as an image in~\cite{bhodia2019transfer, taheri2018leveraging}. 
By bringing pre-trained Convolutional Neural Network (CNN) model that is suitable for image data, the authors do transfer learning to adapt network traffic data~\cite{bhodia2019transfer, taheri2018leveraging}.
Even though this approach is effective, the manners are quite different from the proposed method in that they use pre-trained parameters on image dataset.
Singla et al. propose transfer learning framework using Deep Neural Network (DNN), but this approach requires labeled datasets for both source and target domains contrary to the proposed method not requiring labeled dataset for a target domain\cite{singla2019overcoming}. Moreover, the datasets that are used as source domains and target domains should be generated on the same environment, which can be constraints.

\section{Proposed Model}

In this section, we first present our botnet detection method, which consists of three steps: data pre-processing, anomaly scoring, and anomaly detection.
We then introduce our transfer learning mechanism for botnet detection, with/without the use of labeled data set in the target domain. 

\subsection{Botnet Detection Method}\label{bdm}
\begin{figure}[tb]
    \centering
    \includegraphics[width=0.9\columnwidth]{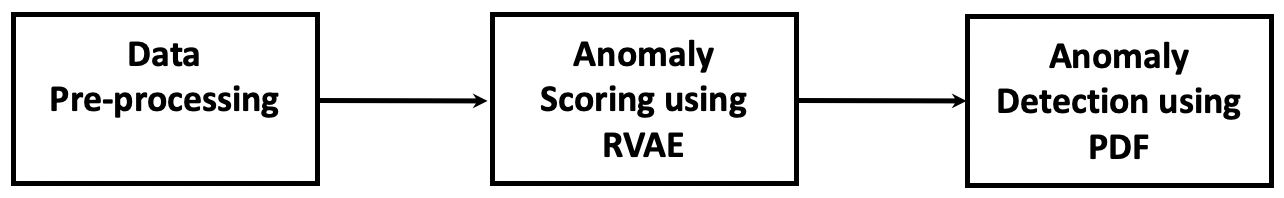}
    \caption{Overview of the proposed botnet detection procedure.}
    \label{fig:proc}
\end{figure}

In order to identify botnets, we propose a novel flow-based botnet detection system coping with sequential nature of traffic flows. 
The overall procedure is shown in Fig.~\ref{fig:proc}, which consists of three steps as follows:

\begin{itemize}
    \item {\bf Data pre-processing:} 
    The data instances are grouped based on a predefined time interval (e.g., 60-second time window), and they are aggregated by host IP addresses.  This process also calculates statistical features and normalizes numerical values.
    \item {\bf Anomaly scoring:}
    At every time window, anomaly scores of every aggregated flow are calculated, which provides a measure of deviation from normal for each individual connection. 
    It is possible to use many different scoring functions.
    In this  work, we assign scores by comparing the input ($x$) with the output from RVAE ($\tilde{x}$). 
    \item {\bf Anomaly detection:}
    Based on the calculated anomaly scores, the anomaly detection function classifies individual connections into either {\em Malicious} or {\em Non-malicious}.
    In particular, our method does not rely on {\em threshold}; rather, it utilizes a couple of probability density function (PDF) that are estimated by normal and botnet instances in training data, respectively.
\end{itemize}

\begin{figure*}[]
    \centering
    \includegraphics[width=0.9\textwidth]{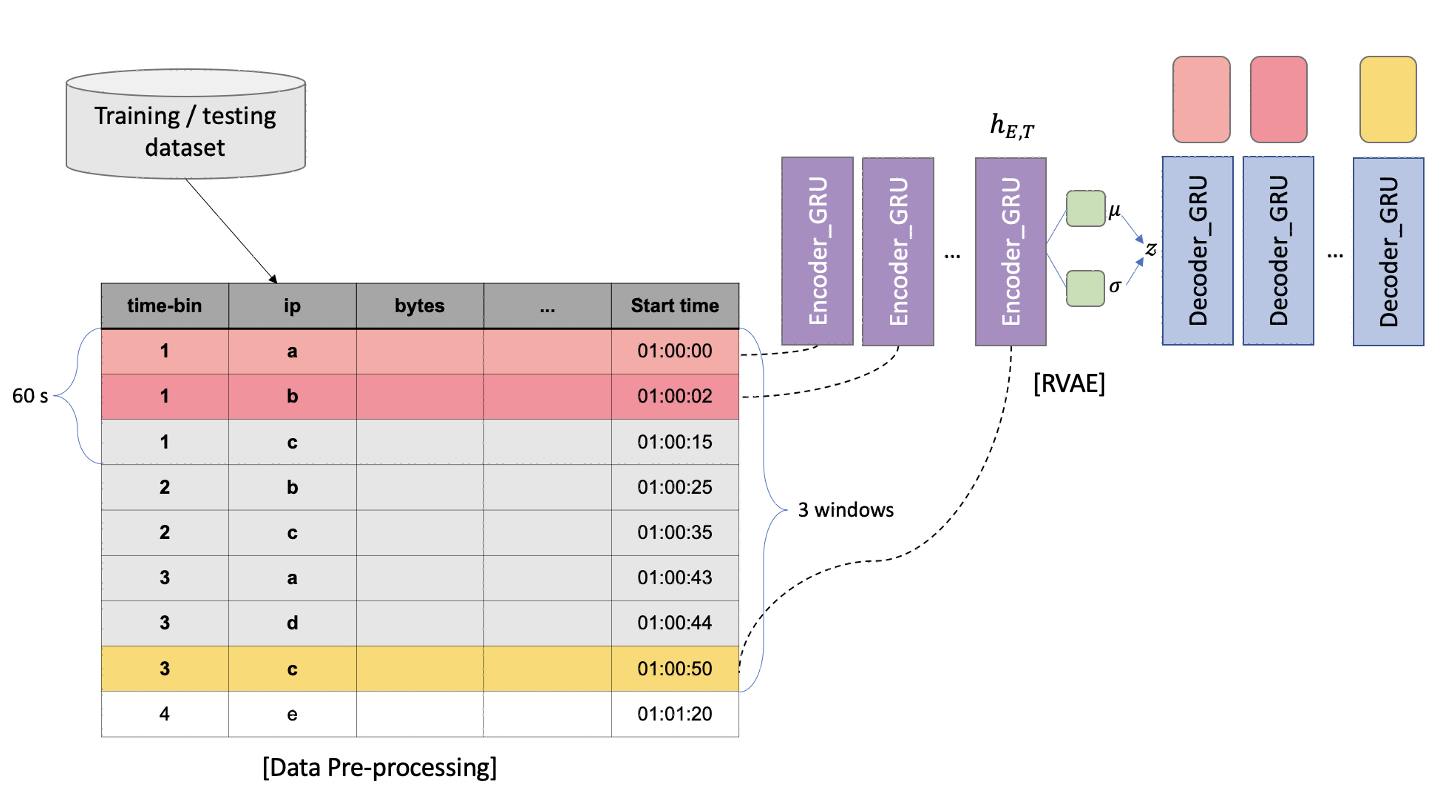}
    \caption{An illustration of out botnet detection procedure using RVAE on a sample of net flows.}
    \label{fig:modelstructure}
\end{figure*}

Fig.~\ref{fig:modelstructure} shows a snapshot of the process for the data pre-processing and anomaly scoring. In the phase of data processing, every flow sorted in chronological order is aggregated to obtain statistic features within the windows. These flow-based features are used as input to RVAE, and are input in the order of time. In the botnet detection system, the encoder distills the common characteristics within the sequential data into latent variable $z$.  The decoder reconstructs sequential inputs from $z$. 
In the end, reconstruction loss is used as the anomaly scoring.

\subsubsection{Anomaly Scores from Recurrent Variational Autoencoder}
\begin{table}[tb]
\caption{Notations used}
\label{tbnotation}
\begin{center}
\begin{tabular}{|l|l|}
\hline
Notation & Description\\
\hline

$h_{E,T}$  & The last hidden state of the encoder\\
$W_\mu$ & Linear transformation to get $\mu(x)$ \\
$W_\sigma$ & Linear transformation to get $\sigma(x)$ \\
$z$ & The latent variable\\
$h_{D,2}$ & The second hidden state of the decoder\\
$h_{D,t}$ & The hidden state of the decoder \\
&at timestep $t$\\
$W^{hh}$ & Linear transformation from the previous \\
&hidden state\\
$W^{hx}$ & Linear transformation from input \\
$W^{s}$ & Linear transformation from hidden state \\
&to get output \\
$\theta$ & The parameters of encoder \\
$x_1$  & The first input of the decoder\\
$\phi$ &The parameters of decoder\\
$\tilde{y_t}$ & Output, reconstruction at timestep $t$\\
${y_{t_n}}$ & $n$th feature within data at timestep $t$\\
$\tilde{y}_{t_n}$ & $n$th feature of reconstruction at timestep $t$\\
$D_{KL}$ & Kullback-Leibler divergence\\
\hline
\end{tabular}
\end{center}
\end{table}
Notations used in this paper are summarized in Table~\ref{tbnotation}.
We first input network traffic data, which are pre-processed, to GRU structure. 
To obtain mean and variance of Gaussian distribution, we use $h_{E,T}$ multiplied by trainable $W_\mu$ and $W_\sigma$, respectively.
With $\mu$ and $\sigma$, $z$ can be obtained, and the $z$ is used as initial hidden state for the decoder. 
\begin{eqnarray}
    &&\mu(x) = W_\mu h_{E,T}\\
    &&\sigma(x) = W_\sigma h_{E,T}\\
    &&z = \mu(x) + \sigma(x)* \epsilon, \epsilon \sim N(0,1)
\end{eqnarray}

\noindent The second hidden state of decoder follows as:

\begin{equation}
     h_{D,2} = sigmoid(W^{hh}z + W^{hx}x_{1})
\end{equation}
\noindent The first input of the decoder($x_1$) is zero-padded.
Finally, the outputs we obtain from RVAE is formulated:
\begin{equation}
     \tilde{y_t} = sigmoid(W^{s}h_{D,t})
\end{equation}
We train $W$ to get minimized the loss function which is defined in \ref{eq:1}.
We train the model with only {\em non-malicious} instances, and in evaluation phase, we calculate reconstruction errors and use it as anomaly scores using both {\em non-malicious} and {\em malicious} instances. 
As we use binary cross entropy as error function, the anomaly score is formulated:
\begin{equation}
    L = \sum_{n=1}^{N} {(1-y_{t_n}) log (1-\tilde{{y}}_{t_n}) + y_{t_n} log \tilde{{y}}_{t_n}}
\end{equation} 
Each time window, we can obtain the anomaly scores of every aggregated flow that belong to the time window. 
In other words, if the traffic connection can be considered malicious or not is indicated by the outputs($L$) of the anomaly detection system. In the following section, we present how to detect botnets with anomaly scores. 

\subsubsection{Anomaly Detection}\label{da}
In many studies, threshold of anomaly scores is used to distinguish whether the source IP addresses in the time window is malicious or not in anomaly detection methods~\cite{nguyen2019gee, dargenio2018exploring, an2015variational}.
The threshold can be set in many ways.  It is a simple and intuitive method; however, some global information about the dataset is required, for example, the ratio of botnets or at least approximate values of anomaly scores of botnet samples. 
Unfortunately, such global information about the traffic data is not known in advance. 
Therefore, this approach can not be used for on-line anomaly detection.

Instead, we suggest a more efficient method using the estimated probability distribution of reconstruction errors, which enables the method be applied in an on-line manner.
In training phase, we collect reconstruction errors from normal and abnormal instances. Then, we find the distribution and its parameters to represent the distribution of reconstruction errors of abnormal and normal, respectively, by exploring various types of distributions and selecting the minimum sum of squared estimate errors (SSE). We call the distribution with the smallest SSE as the {\em best-fit PDF}. We search for the {\em best-fit PDF} among many different candidates such as {\em gamma} distribution, {\em generalized} {\em logistic} distribution, {\em fold} {\em cauchy} distribution, {\em Mielke} distribution and {\em beta} distribution, among others. In testing phase, the estimated PDF can be utilized to obtain likelihood to belong to each distribution. Comparing the likelihood values of the two different distributions, we assume that each sample of the test data set belongs to the distribution showing the higher likelihood. 
Using {\em best-fit PDFs} does not requrie global information about the data set and could be used for on-line botnet detection.

\subsection{Transfer Learning for Botnet Detection}
We further propose a novel transfer learning technique for botnet detection.
Unlike the semi-supervised RVAE method we present in the previous section \ref{bdm} requiring labeled data on the domain of interest, the training framework lets the botnet detection system perform well on partially labeled or even unlabeled datasets in the target domains.
Both the ML structure that produces anomaly scores and the anomaly detection method utilizing reconstruction errors maintain in the transfer learning framework.
The only different thing is the way to train the RVAE: each minibatch sample from a source and a target domain is used for training, respectively and sequentially.

We follow the procedure of transfer anomaly detection method proposed in~\cite{kumagai2019transfer}, and adapt it to the characteristics of network traffic data.  Since it is hard to obtain labeled data for training, we further develop the method to be trained without label information on the target domain.
Namely, we consider two cases of training data on botnet detection:labeled dataset on the target domain ($with\_label$) and unlabeled dataset on the target domain ($without\_label$).
In the both methods, normal and anomalous instances in a source domain are used for training RVAE at first.
After calculating the gradients with a minibatch of samples of the source domain and updating the parameters of the decoder and the encoder, we use a minibatch of samples of the target domain for training. Then, we calculate and update the gradients in the same manner.

$X^{+}_s$ is a set of anomalous instances in a source domain ($x^+_s \in X^{+}_s$).
$X^{-}_s$ is a set of normal instances in a source domain ($x^-_s \in X^{-}_s$).
$X^{+}_t$ is a set of anomalous instances in a target domain ($x^+_t \in X^{+}_t$).
$X^{-}_t$ is a set of normal instances in a target domain ($x^-_t \in X^{-}_t$).
$N^+_s$, $N^-_s$ is the number of instances of anomalous and normal on the source domain.

We use the objective function of the source domain in~\cite{kumagai2019transfer}:
\begin{equation}
    s_{\phi,\theta}(x|z) = \sum_{n=1}^{N} {(1-x_{n}) log (1-\tilde{{x}}_{n}) + x_{n} log \tilde{{x}}_{n}}\\
\end{equation}

\begin{eqnarray}
\label{eqn:source_loss_recon}
    L_s(\theta,\phi|z) &=& \frac{1}{N_s^{-}}\sum^{N_s^{-}}_{n=1} {s_{\phi,\theta}(x^-_s|z)} \\ \nonumber && -\frac{\lambda}{N_s^-N_s^+}\sum_{n,m=1}^{N_s^-,N_s^+}{f(s_{\phi,\theta}(x_s^+|z)-s_{\phi,\theta}(x_s^-|z))}
\end{eqnarray}

\begin{eqnarray}
\label{eqn:source_loss}
    \mathbb{L}_s(\phi, \theta) &=& \mathbb{E}_{q_\phi(z|x)}[L_s(\theta,\phi|z)] \\ \nonumber && + \beta D_{KL}(q_{\phi}(z|X_s^-)|p(z))
\end{eqnarray}

We do not use \textit{latent domain vectors} that is denoted as $z$ in~\cite{kumagai2019transfer} as there is only one domain for each source and target domain.
Since we utilize VAE contrary to AE, $z$ represents the latent variable in the proposed method~\cite{kingma2013auto}.

The proposed method can be categorized into two based on whether the labeled dataset on the target domain is used or not.
The overall process is identical, but transfer learning with the unlabeled data set on the target domain is different from the method using the labeled data set on the target domain. It uses entire instances in the target domain for training.
On the other hand, only normal instances in the target domain are made use of for training on {$with\_label$} method.
Therefore, there is a slight difference in the objective function of the target domain of the two methods, which you can find in the next subsection.
In the case of the source domain, the objective functions on both methods are equal to  Equation~\ref{eqn:source_loss}.

\subsubsection{Using labeled data set in a target domain}
In this case, we use only normal instances for training likewise other semi-supervised learning methods~\cite{nicolau2018learning,dargenio2018exploring}.
The difference from the semi-supervised learning is that we utilize transferred knowledge from the source domain in order to classify instances of the target domain.  Therefore, we can expect better results than semi-supervised learning methods. Furthermore, even if there is not enough labeled data, it might be possible to obtain comparable performance to other semi-supervised learning methods.
The objective function for the target domain in {\em with\_label} method is formulated as in~\cite{kumagai2019transfer}:
\begin{eqnarray}
\label{eqn:with_label_loss}
    \mathbb{L}_t(\phi, \theta) &=& \mathbb{E}_{q_\phi(z|x)}\left[\frac{1}{N_t^{-}}\sum^{N_t^{-}}_{n=1} {s_{\phi,\theta}(x^-_t|z)}\right]\\ \nonumber && + \beta D_{KL}(q_{\phi}(z|X_t^-)|p(z))
\end{eqnarray}
The overall process of the method is shown in Algorithm~\ref{alg:generator}. 
As you can see in Algorithm~\ref{alg:generator}, we first utilize instances on the source domain. For training the model on the source domain, both abnormal and normal samples are required to compute the loss function using Equation~\ref{eqn:source_loss_recon}.
After updating the parameters of the model with the source domain dataset, we sample minibatch of the normal instances in the target domain, and then update the parameters.
We iterate the process until all instances are utilized for training.
Note that this method requires normal instances to be labeled in the target domain.


\begin{algorithm}[tb]
\SetAlgoLined
\caption{The Procedure of Training Transfer Anomaly Detection $with\_label$ Method}
\label{alg:generator}
\textbf{Input}: instances of source domain $x^{-,+}_s \in X^{-,+}_s$ and instances of target domain $x^{-}_t \in X^{-}_t$\\
\textbf{Output}: $G_\theta, F_\phi$\\
\SetKwProg{generate}{Procedure }{}{end}
\For{number of epochs}{
Sample minibatches from $X^{+}_s$, $X^{-}_s$ and $ X^{-}_t$\\
($B_{x^{-}_s}$ $\subset$ $X^{-}_s$,
$B_{x^{+}_s} \subset X^{+}_s$,
$B_{x^{-}_t}$ $\subset$ $X^{-}_t$)\\
\For{$B_{x^{-}_s}^a, B_{x^{+}_s}^c, B_{x^{-}_t}^e,(a=1,...,A,c=1,...,C, e=1,...,E$)}{
     \ForAll{$x^{-}_s \in B_{x^{-}_s}^a$}{
        $\tilde{{x}}_{s}^-$ = $G_{\theta}$($F_{\phi}(x^{-}_s$))\\
        \ForAll{$x^{+}_s \in B_{x^{-}_s}^c$}{
            $\tilde{{x}}_{s}^+$ = $G_{\theta}(F_{\phi}(x^{+}_s$)) \\
        }
        Update the Encoder and the Decoder by descending its stochastic gradient:
        $\bigtriangledown_{\theta,\phi}(L_s(\phi,\theta))$\\
     }
     \ForAll {$x^{-}_t$ $\in$ $B_{x^{-}_t}^e$}{
        $\tilde{{x}}_{t}^-$ = $G_{\theta}(F_{\phi}(x^{-}_t$))\\
     }
     Update the Encoder and the Decoder by descending its stochastic gradient:
      $\bigtriangledown_{\theta,\phi}(L_t(\phi,\theta))$\\
     }
}
\end{algorithm}

\subsubsection{using unlabeled data set in a target domain}
In this method, we assume the situation where there is no labeled data set on the target domain. 
That means we cannot distinguish between normal and abnormal examples.
To deal with such a case, we use an entire instance of the dataset in the target domain only for the first several epochs ($E$).
From the very next sequence after $E$ epochs, we collect instances that show lower reconstruction errors in each minibatch. 
Given characteristics of datasets in that the number of normal samples is much higher than the number of anomalies, it is inferred that the lower instance of reconstruction errors is more likely to be normal.
In order to give weight to the estimated normal instances, we use the instances more than once in the following minibatch training.

To do this, we sort the instances by the size of reconstruction errors every minibatch.
We then select instances of the bottom $r\%$ of reconstruction errors in minibatch as the estimated normal samples and add the instances to the following minibatch training samples. 
In other words, the sample used in the next step of training consists of the next step minibatch and the part of the previous samples with lower reconstruction error.
By selecting samples this way, we can train the anomaly detector effectively even with an unlabeled dataset on the target domain.

$M_t$ is the number of the increased samples due to selection, and varies depending on the ratio($r$).
$w_{x_t}$ is weight on each instance now that instances with lower reconstruction errors are used more than once.
The objective function for target domain in {\em without\_label} method is formulated:

\begin{eqnarray}
\label{eqn:without_label_loss}
    \mathbb{L}_t(\phi, \theta) &=& \mathbb{E}_{q_\phi(z|x)}\left[\frac{w_{x_t}}{M_t}\sum^{M_t}_{n=1} {s_{\phi,\theta}(x_t|z)}\right] \\ \nonumber && + \beta D_{KL}(q_{\phi}(z|X_t)|p(z))
\end{eqnarray}

\begin{algorithm}[tb]
\SetAlgoLined
\caption{The Procedure of Training Transfer Anomaly Detection $without\_label$ Method}
\label{alg:with_label}
\textbf{Input}: instances of source domain $x^{-,+}_s \in X^{-,+}_s$ and instances of target domain $x^{-}_t \in X_t$\\
\textbf{Output}: $G_\theta, F_\phi$\\
\SetKwProg{generate}{Procedure }{}{end}
\For{the number of epochs}{
Sample minibatches from $X^{+}_s$, $X^{-}_s$ and $ X_t$\\
($B_{x^{-}_s}$ $\subset$ $X^{-}_s$,
$B_{x^{+}_s} \subset X^{+}_s$,
$B_{x_t}$ $\subset$ $X_t$)\\
\For{$B_{x^{-}_s}^a, B_{x^{+}_s}^c, B_{x_t}^e,(a=1,...,A,c=1,...,C, e=1,...,E$)}{
     \ForAll{$x^{-}_s \in B_{x^{-}_s}^a$}{
     $\tilde{{x}}_{s}^-$ = $G_{\theta}$($F_{\phi}(x^{-}_s$))\\
     \ForAll{$x^{+}_s \in B_{x^{-}_s}^c$}
     {
     $\tilde{{x}}_{s}^+$ = $G_{\theta}(F_{\phi}(x^{+}_s$))\\
     }
      Update the Encoder and the Decoder by descending its stochastic gradient:
      $\bigtriangledown_{\theta,\phi}(L_s(\phi,\theta))$\\
     }
     \ForAll {$x_t$ $\in$ $B_{x_t}^e$ and the previous samples}
     {
     $\tilde{{x}}_{t}$ = $G_{\theta}(F_{\phi}(x_t$))\\
     }
     Update the Encoder and the Decoder by descending its stochastic gradient:
      $\bigtriangledown_{\theta,\phi}(L_t(\phi,\theta))$\\
      Add instances of the bottom $r\%$ reconstruction errors to the following minibatch training samples. 
     }
}
\end{algorithm}

\section{Experiments}\label{sec:exp}
We have experimented several ways to validate reliability of the proposed method in different aspects. 
The first is to show that the proposed RVAE structure has better performance than both Random Forest and the standard VAE, which we call as MLP-VAE in this paper.
Second, we explain how the reconstruction errors are distributed and how to utilize it in detecting botnets.
Third, we demonstrate the effectiveness of the transfer learning framework using two different network traffic datasets.

We use the 5 different evaluation metrics to validate our performance when we show compare the RVAE with the other semi-supervised learning method and the supervised learning method; Area Under the Receiver Operating Characteristics (AUROC), Area Under the Precision-Recall Curve (AUPRC), Precision, Recall and F1 score. We present the results from the model showing the best value of AUPRC in 5-fold cross validation sets.
When we evaluate the transfer learning framework, we use AUROC, True Positive Rate (TPR), False Positive Rate (FPR), True Negative Rate (TNR) and False Negative Rate (FNR). In this case, we present the results from the model showing the best value of AUROC in validation set to avoid over-fitting. We use the mean of outputs each metric on the five identical experiments.
The source code is written with the PyTorch\footnote{https://pytorch.org} library.

\subsection{Evaluation Datasets}

We utilize two different datasets to test the performance of the proposed model.

\subsubsection{Dataset used for Evaluating RVAE}\label{B}
We use CTU-13 dataset widely used in the studies of botnet detection~\cite{ nguyen2019gee,  nicolau2018learning, dargenio2018exploring, an2015variational, torres2016analysis, du2019fenet, ongun2019designing, garcia2014empirical}.
In this dataset, a botnet scenario is a particular infection of the virtual machines using a specific malware. Thirteen of these scenarios were created, and each of them was designed to be representative of some malware behavior~\cite{garcia2014empirical}.
To compare the results of MLP-VAE and Random Forest, we reproduced nearly the same experimental settings reported in \cite{nguyen2019gee} and \cite{ongun2019designing}.
In~\cite{nguyen2019gee} and~\cite{ongun2019designing} that proposes VAE and Random Forest structures respectively that we select as the baseline, they prove the robustness of their methods on scenario 1, 2 and 9 of CTU-13 dataset, which consists of only botnet called Neris. The Neris is the IRC based bot infecting other machines by Spam and Click Fraud. In our experiments, all methods show the similar performance in every metrics, as shown in Table~\ref{table3}. Especially, Random Forest performs very well on the testing dataset because botnet families in the testing dataset are already used for training. In other words, Random Forest method is able to capture dominant features to classify anomalies.

To test the cases where the training data might be different from testing data, we plan to separate the CTU-13 dataset as suggested by~\cite{garcia2014empirical}.
The idea is to separate the data set so that none of the botnet families used in the training and cross-validation datasets should be used in the testing dataset. 
A method that achieves high accuracy in this case is likely to detect new behaviors. 
By splitting of CTU-13 data this way, we also mimic the real situation where the operations of botnet changes over time in terms of protocols and attack types. 
Compared to the restricted dataset (scenario 1, 2, 9), various types of botnets that have IRC-based, P2P-based and HTTP-based communication methods and conduct attacks such as Spam, Click Fraud, Port Scan, DDos and FastFlux are included in the dataset that we use for the experiments.
The datasets describing which scenarios are included are in Table~\ref{table2}.

\begin{table}[tb]
\caption{CTU-13 Dataset}
\label{table2}
\begin{center}
\begin{tabular}{|c|c|c|c|c|}
\hline
Dataset&Scenario\\
\hline
Training\&Validation&3,4,5,7,10,11,12,13\\
\hline
Testing &1,2,6,8,9\\
\hline
\end{tabular}
\end{center}
\end{table}

\begin{table*}[tb]
\caption{Results comparison : Trained and tested on different dataset}
\label{table3}
\begin{center}
\begin{tabular}{|c|c|c|c|c|c|c|c|}
\hline
Dataset&Model&Recall&Precision&F1&AUPRC&AUROC\\
\hline
&RVAE&0.964&0.914&0.938&0.966&0.971\\
Scenario 1,2 and 9&MLP-VAE&0.963&0.922&0.941&0.954&0.965\\
&Random Forest&\textbf{0.996}&\textbf{1.000}&\textbf{0.998}&\textbf{0.999}&\textbf{0.999}\\
\hline
&RVAE&\textbf{0.970}&0.871&0.918&\textbf{0.976}&\textbf{0.979}\\
Scenario 1-13&MLP-VAE&0.941&0.936&\textbf{0.938}&0.967&0.963\\
&Random Forest&0.696&\textbf{0.997}&0.820&0.921&0.936\\
\hline
\end{tabular}
\end{center}
\end{table*}

\subsubsection{Dataset used for Evaluating Transfer Learning Framework}
In transfer learning framework, data are required in source domains and target domains, respectively.
The existing studies on transfer learning usually use the same dataset for target domains and source domains~\cite{kumagai2019transfer, alothman2018similarity, taheri2018leveraging, bhodia2019transfer, singla2019overcoming, jiang2019new}.
However, we want to detect suspicious botnet connections on a new network monitoring dataset (target domain). 
In this case, we cannot help using the target domain that is different from the source domain since the data on the target domain cannot be labeled.
Therefore, we need two different datasets to test transfer learning framework.

For the dataset in the source domain, we use CTU-13 dataset that is used to test RVAE model.
CTU-13 dataset is labeled network traffic data.
In this experiment, we only focus on botnet called Neris, which is used in scenario 1,2 and 9 in CTU-13 dataset, to reduce complications.
We use whole data instances rather than separating $Normal$ labels from $background$ labels.
The scenario 1,2 and 9 were collected for three days. 
On the other hand, for the dataset in the target domain, we use a network monitoring data set from a large research institute(called dataset K).
The dataset is collected using a Zeek server connected at the network border.
The Zeek server was installed all-in-one and used a default policy.
We use the data collected for one day among seven days to balance the size of the target domain data with the source domain data.
The network monitoring data is not labeled.

In order to use transfer learning framework, each dataset in the source/target domain should be related.
Therefore, even though every study that uses CTU-13 dataset 
utilizes \textit{Netflow} type of data that is provided in~\cite{garcia2014empirical}, we utilize CTU-13 data from Zeek software as dataset K is obtained from Zeek as well. 
In our experimental setting, the source domains and the target domain are different but related as we utilize two dataset generated on different environment but collected from Zeek.
To sum up, we utilize CTU-13 Zeek dataset as a source domain, and use dataset K as a target domain.


\subsection{Labeling method}
Both dataset K and CTU-13 Zeek data have no labels.
To quantify the accuracy of transfer learning, we need a way to assign labels to these data sets.
We present the new labeling method that can be applied to both CTU-13 and dataset K.
Zeek's event engine records weird activities that can indicate malformed connections, malfunctioning or misconfigured hardware, or an attacker attempting to avoid/confuse a sensor. 
Also, Zeek provides specified type indication the reason why the connections provoke weird flags.
Those suspicious connection are logged in a file named \textit{weird.log}.
However, we find that \textit{weird.log}  has no clear correlation with botnet label in CTU-13 \textit{Netflow}  dataset. 
Many of the connections logged in \textit{weird.log}  might be made by misconfigured hardware or malformed connections, not related to botnet. 
Most connections that are made by botnet in \textit{Netflow}  are even not detected as "weird" activity in Zeek.

On closer examination, we find that botnet called Neris accounts for 84\% of connections representing {\em irc\_line\_too\_short} in \textit{weird.log}  among data from 13 scenarios. 
In addition, Neris accounts for 82\% of connections representing {\em irc\_invalid\_line} among data from 13 scenarios. 
We infer that most connections in the \textit{weird.log}  representing {\em irc\_line\_too\_short} and {\em irc\_invalid\_line} are given by Neris.

Therefore, we decide to use the indication information from \textit{weird.log}, and label the host IP address with {\em irc\_line\_too\_short} and {\em irc\_invalid\_line} as {\em malicious}. 
With the collected host IP addresses that are {\em malicious}, we can use network log features from \linebreak  \textit{conn.log}  composed of source/destination IP addresses, ports, time, protocol, duration, number of packets, number of bytes, state, and service.
As both CTU-13 Zeek dataset and dataset K are produced by Zeek, the labeling method can be applied commonly to the two different datasets.

\subsection{Data Pre-processing}\label{sec:data}
We use the same data pre-processing method for CTU-13 dataset and dataset K. The data are composed of source and destination IP addresses and ports, time, protocol, duration, number of packets, number of bytes, state, and service. We process the data to use the aggregated flows statistic, which is the way many existing works adopt in order to obtain flow-based features~\cite{nguyen2019gee, ongun2019designing, nicolau2018learning, dargenio2018exploring, an2015variational, torres2016analysis}. 
For the data collected from Zeek, we add to use missed bytes and include more types of service such as \textit{mysql, imap} and \textit{ftp} compared to the original CTU-13 dataset. 
We group the connections at every time interval of $T$, and aggregate features within every group based on the source IP addresses to generate flow-based features.

With the processing method, we can detect IP address showing the malicious behavior in a particular time window. 
Many existing works experimentally find the most appropriate time window $T$, which is crucial, while too small time window might not capture traffic characteristics over a longer period of time, too large time window cannot provide timely detection in waiting for the end of the window~\cite{zhao2013botnet, nguyen2019gee, garcia2014empirical, ongun2019designing, dargenio2018exploring, an2015variational, torres2016analysis}.
We experimented changing the duration of windows to find the ideal value for the statistical aggregation.
We then sort the entire data within the time window by the time of the source IP connection group, because the RNN model is sensitive to the order of the inputs. 
For RVAE, we use the network traffic connections collected within $N$ windows as the sequential inputs to the model. 
You can see it in the data processing part of Fig.~\ref{fig:modelstructure}. 
In the case of Fig.~\ref{fig:modelstructure}, 60-second duration of three windows are used. 

In terms of source/destination ports and destination IP addresses, we count the number of unique records with connected source IP addresses in the time window. 
In addition, for the source IP addresses, we count the number of connections with the source IP addresses in the time window. 
For service, state, and protocol, we count the number of different values in each category with the source IP addresses in the time window. 
Finally, we normalize the numerical values to be between 0 and 1 using min-max normalization. 
As a result, the number of features used in this experiment is smaller compared to the number of features used in~\cite{ongun2019designing} and~\cite{nguyen2019gee}.

\begin{figure*}[tb]
\centering 
\subfloat[duration of window : 5s]{%
  \includegraphics[width=0.33\linewidth]{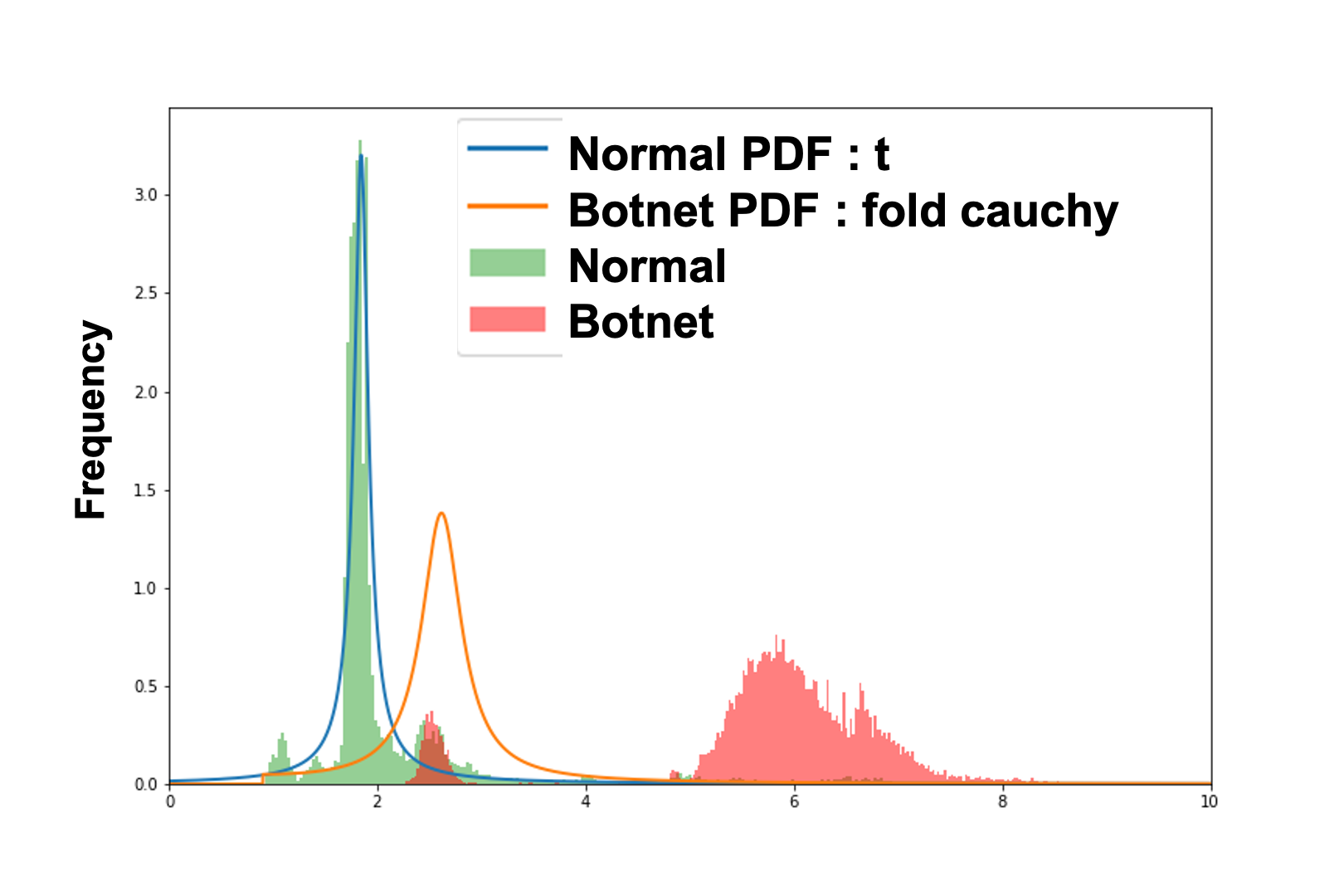}
  \label{fig:recon_error_5}
}
\subfloat[duration of window : 60s]{%
  \includegraphics[width=0.33\linewidth]{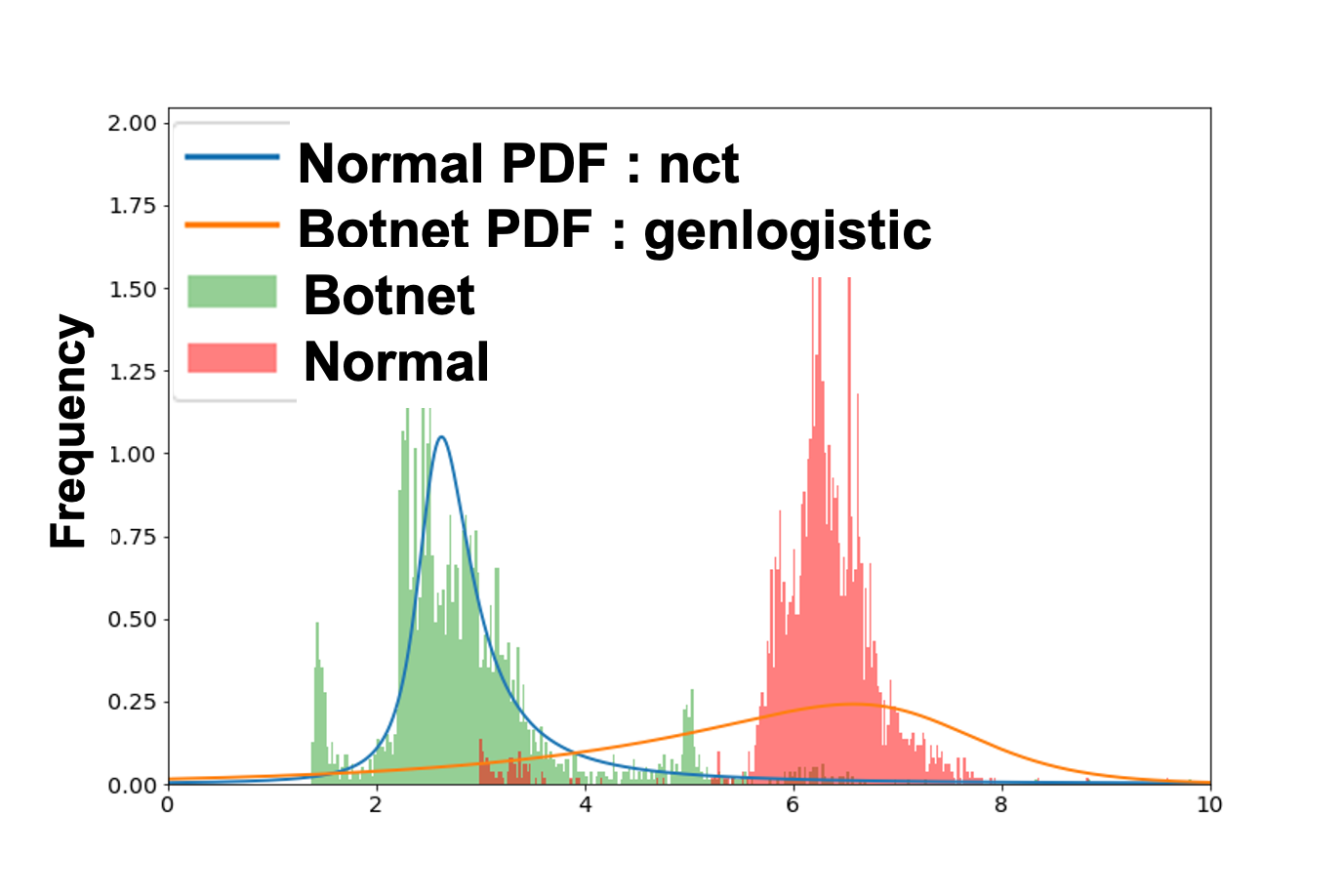}
  \label{fig:recon_error_60}
}
\subfloat[duration of window : 300s]{%
  \includegraphics[width=0.33\linewidth]{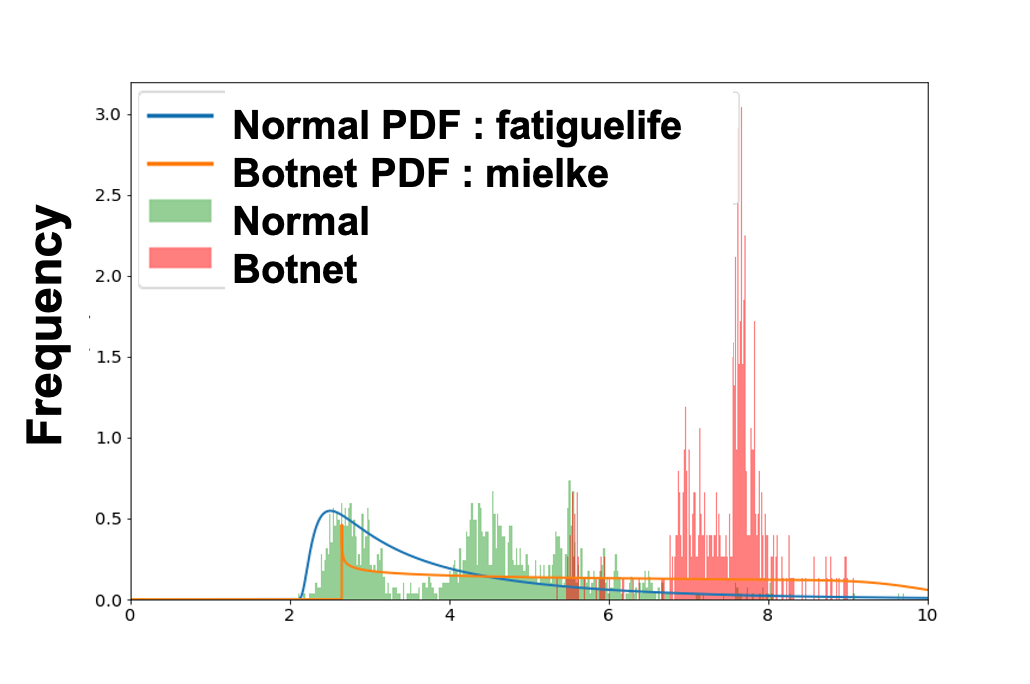}
  \label{fig:recon_error_300}
}
\caption{Distribution of reconstruction errors of {\em normal} and {\em botnet} samples with different duration of window}
\label{fig:recon}
\end{figure*}

\subsection{Experimental Configuration}
The architecture of MLP-VAE, which is used as an anomaly detector, follows the one from ~\cite{nguyen2019gee}, [\# of features $\to$ 512 $\to$ 512 $\to$ 1024 $\to$ 100].
For RNN architecture, we use 2-layer bidirectional GRU. We use the 512 hidden states, and 100  latent variable as MLP-VAE. 
We also apply {\em ReLU} activation~\cite{nair2010rectified} to MLP-VAE as well as RVAE. 
The Kullback-Leibler annealing method is set so that the weight ($\beta$) multiplied to KLD increases linearly for 500 gradient updates for RVAE. 
We train for 500 epochs with Adam optimizer and 128 batch-size.
Also, the learning rate is set as 0.01 for both VAE models.

\subsection{Comparison Methods}
There are two experiments in this paper. First, we compare the experimental results of the proposed model, RVAE, to the results of MLP-VAE and Random Forest. In terms of MLP-VAE, the experimental results are based on the same data processing method with the same optimizer, learning rate and the size of latent variable from our experiment. 
Second, in order to evaluate that the anomaly detector performs well on the dataset K, we compared the transfer learning results to the semi-supervised learning. For both cases, RVAE structure is used as an anomaly detector.

\section{Results and Discussion}
We now report our experimental results. We first discuss the performance of anomaly detection based on RVAE semi-supervised learning with other methods using a set of different metrics.
Then we  compare  the  performance  of  RVAE  transfer  learning method  with  RVAE  semi-supervised  learning  on  different measures.

\begin{table*}[tb]
\caption{Results comparison on different scenarios}
\label{table4}
\begin{center}
\begin{tabular}{|c|c|c|c|c|c|c|c|}
\hline
Dataset&Model&Recall&Precision&F1&AUPRC&AUROC\\
\hline
\multirow{2}{*}{Scenario 1}&RVAE&\textbf{0.979}&0.709&0.826&\textbf{0.888}&\textbf{0.969}\\
&MLP-VAE&0.968&\textbf{0.729}&\textbf{0.832}&0.729&0.938\\
\hline
\multirow{2}{*}{Scenario 2}&RVAE&\textbf{0.941}&0.970&\textbf{0.955}&0.967&\textbf{0.967}\\
&MLP-VAE&0.936&\textbf{0.974}&\textbf{0.955}&\textbf{0.968}&0.964\\
\hline
\multirow{2}{*}{Scenario 6}&RVAE&\textbf{0.943}&0.612&\textbf{0.742}&\textbf{0.728}&\textbf{0.867}\\
&MLP-VAE&0.689&\textbf{0.646}&0.666&0.679&0.769\\
\hline
\multirow{2}{*}{Scenario 8}&RVAE&\textbf{0.973}&0.937&\textbf{0.954}&\textbf{0.980}&\textbf{0.981}\\
&MLP-VAE&0.921&\textbf{0.974}&0.947&0.971&0.959\\
\hline
\multirow{2}{*}{Scenario 9}&RVAE&0.957&\textbf{0.989}&\textbf{0.973}&\textbf{0.991}&\textbf{0.976}\\
&MLP-VAE&\textbf{0.961}&0.986&\textbf{0.973}&0.990&\textbf{0.976}\\
\hline
\end{tabular}
\end{center}
\end{table*}

\subsection{RVAE semi-supervised learning}\label{ssl}
For a quantitative evaluation, we compare the performance of our method with others on different metrics. 
For qualitative evaluation, we plot the distribution of the reconstruction errors of the normal and botnet cases. 
Moreover, we plot estimated the {\em best-fit PDF} described in the section~\ref{da}. 
To compare methods with the same processing and detection method, we reproduce MLP-VAE and RF in~\cite{nguyen2019gee} and~\cite{ongun2019designing}, respectively.  
While the value in the literature~\cite{nguyen2019gee} is 0.936, our reproduced value of AUROC with MLP-VAE is 0.963. 
Even if there may be a slightly different experimental setup, the reproduced results show that our implementation is still valid according to a bit higher result than the result of the original paper.

\noindent\textbf{Performance comparison among methods.} 
In Table~\ref{table3}, the RF method shows the nearly perfect performance in every metric, even though VAE models show the comparable performance. 
This is because the training/testing datasets are based on scenario 1, 2 and 9 share the same characteristics. 
RF is effective in finding dominant features in these restricted datasets. 
However, as we mention in the section~\ref{B}, we want to evaluate the cases where the training and testing data sets are different. 
In Table~\ref{table3}, we show the results from the training and testing on the whole dataset that we mentioned in the section~\ref{B}. In this experiment, we pre-processed our data by using 60-second duration of window and using three windows.
While both restricted training and testing dataset(Scenario 1,2 and 9) consist of only Neris, the entire testing and training dataset(Scenario 1-13) consist of each different botnets: Rbot, Virut, Sogou, and NESIS.ay are used for training, and Neris, Menti, and Murio are used for testing. 
Because each botnet shows different characteristics, there is an overall performance degrade with RF that is affected by the dominant features of the training dataset. 
Nonetheless, VAE methods validate its reliability by showing the robust performance with the generalized dataset. 
In addition, we find that RVAE method outperforms MLP-VAE method overall based on the same features and the same size of latent variables on both datasets, as you see in Table~\ref{table3}.
It can be concluded that the botnets of network traffic flow data should be detected utilizing sequential and periodic patterns.

\begin{table*}[tb]
\caption{Results comparison : different window duration}
\label{table6}
\begin{center}
\begin{tabular}{|c|c|c|c|c|c|c|}
\hline
Window&\multirow{2}{*}{Recall}&\multirow{2}{*}{Precision}&\multirow{2}{*}{F1}&\multirow{2}{*}{AUPRC}&\multirow{2}{*}{AUROC}\\
duration(s)&&&&&\\
\hline
5&0.852&0.912&0.881&0.948&0.913\\
60&\textbf{0.970}&\textbf{0.871}&\textbf{0.918}&\textbf{0.976}&\textbf{0.979}\\
300&0.686&0.563&0.619&0.724&0.783\\
\hline
\end{tabular}
\end{center}
\end{table*}

\noindent\textbf{Probability density function of reconstruction errors.}
As shown in Fig.~\ref{fig:recon}, the distribution of the reconstruction errors of botnet samples can be distinguished from the distribution of the normal sample reconstruction errors. 
As we only use {\em non-malicious} samples for training, we expect that the reconstruction errors of {\em malicious} samples are larger than that of the  {\em non-malicious} samples. 
Comparing medians of those two distributions, we can notice that the median of the distribution of {\em non-malicious} reconstruction errors is larger than the median of the distribution of botnet reconstruction errors.

Especially, you can find a group of botnet samples that have the smaller reconstruction errors compared to the other botnet samples in Fig.~\ref{fig:recon_error_60}.
We focus on the samples whose reconstruction errors are smaller than 4. 
We find that 66\% of the samples of the scenario 6 labeled as botnet show the reconstruction errors less than 4, while only 0\%$\sim$4\% of samples in the other scenario show reconstruction errors less than 4.
The scenario 6 utilizes proprietary command control channels unlike other scenarios most of which use IRC, HTTP and P2P communication methods~\cite{garcia2014empirical}.
The samples of the group having small reconstruction errors show low values for DNS, smtp, ssl, the number of IP addresses, the number of ports, and the number of different IP addresses in window. 
These characteristics mainly represent {\em non-malicious} other than the botnet.
We conclude that the general nature of the scenario 6 makes dozens of samples obtain the smaller reconstruction errors. 

\noindent\textbf{Results on each scenario.}
In Table~\ref{table4}, you can find the experiment results tested on each scenario. 
Overall, RVAE shows better performance than MLP-VAE.
What stands out is that there exists a tendency that the MLP-VAE performs better than the RVAE on the precision, and the RVAE performs better than the MLP-VAE on the recall, as you can also find the same aspect in Table \ref{table3}.
While the performance difference between the two methods is not very clear on the botnet called Neris (Scenario 1,2,9), it is clear that the RVAE surpasses the MLP-VAE, especially in Scenario 6.
Although in terms of scenario 6, which consists of botnet called Menti, both RVAE and MLP-VAE shows somehow lower performance than other scenarios, RVAE is much more superior to MLP-VAE compared to other scenarios. 
As we mentioned above, botnet characteristics of scenario 6 are unlikely to be distinguished from normal instances. 
The AUROC of RVAE is 12\% higher than that of MLP-VAE in the scenario 6, while the difference between RVAE and MLP-VAE in others is about 0.0\%$\sim$3.0\%.
We can interpret that in the situation where botnet characteristics of the individual connections are not distinct, the method utilizing sequential properties of input helps the anomaly detector to detect botnet.

\noindent\textbf{Duration of window.}
In order to propose the right duration of window, our experiments have been done with changing the duration of window to 5 seconds, 60 seconds and 300 seconds in Table~\ref{table6}.
In general, the performance of 60-second duration of window are higher than those of other duration lengths.
We infer that as we use a long duration, the number of source IP addresses that belongs to the same time window increases, which aggravates the vanishing gradients problem in a long-term sequence.
On the other hand, too short duration cannot provide efficient length to represent the patterns of the time windows with statistically aggregated values. 
From our experiments, 60-second duration is the most suitable, as you can find in Table~\ref{table6} quantitatively and Fig.~\ref{fig:recon} qualitatively.

\subsection{RVAE transfer learning}
Now that we have validated the effectiveness of using RVAE structure as an botnet detector in \ref{ssl}, we conducted experiments to show that transfer learning framework allow the botnet detector working on the new dataset K.
To present its possibility, we compare the performance of RVAE transfer learning method with RVAE semi-supervised learning on different measures. 

\begin{figure}[!tb]
    \centering
    \includegraphics[width=0.9\columnwidth]{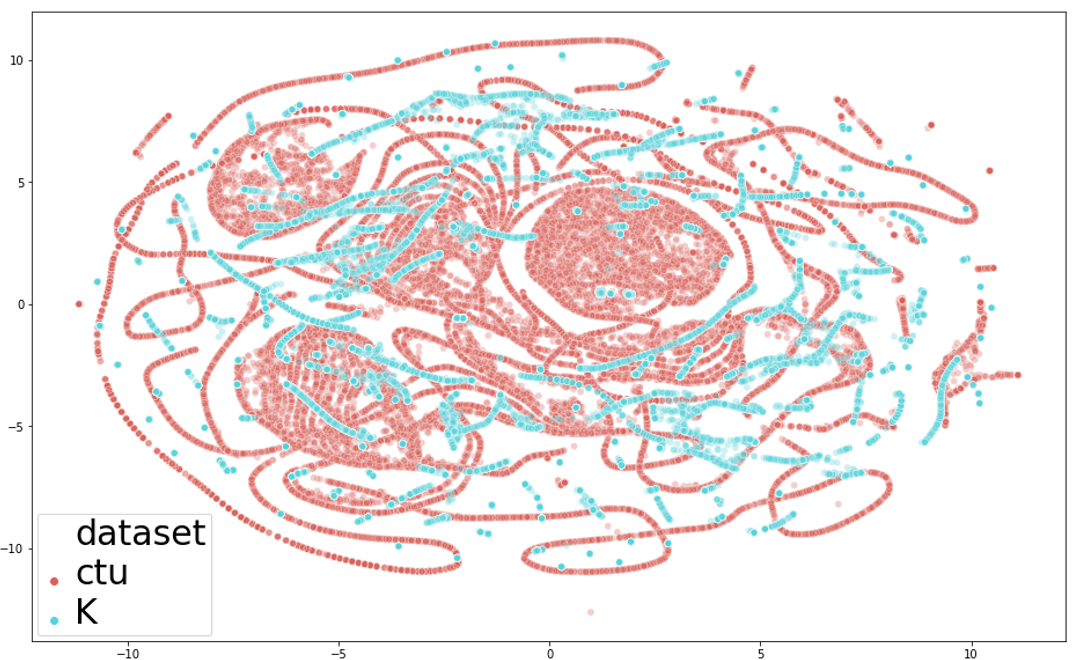}
    \caption{t-SNE plot over the source and the target domains}
    \label{fig:tsne}
\end{figure}

\noindent\textbf{t-SNE plot of each dataset.} We plot t-SNE~\cite{maaten2008visualizing} of each instance of two datasets to justify to use transfer learning.
To use transfer learning, two domains should be related and share common characteristics.
We reduce dimension of features of data from 48 to 2 in order to visualize its distribution.
The source domain dataset and the target domain dataset are not generated in the same environment. 
The source dataset, CTU-13 is made for the purpose of research for botnet detection in the environment where attacks of botnet are controlled.
On the other hand, the target domain dataset K is network monitoring data that is collected using a Zeek server connected to the switch between the Internet and the local network.
Therefore, the two distributions cannot be completely overlapping, as you can see in Fig.~\ref{fig:tsne}.
However, because both data share common characteristics generated from Zeek, the two distributions are not completely separated.
As a result, transfer learning, especially the application of {\em transductive transfer learning}, can show the improved performance over semi-supervised learning.

\begin{table*}[tb]
\caption{Average of each metric over the target domain (dataset K), various $r_s$ (0.05, 0.1, 0.5) in the $without\_label$ method}
\label{main_result}
\begin{center}
\begin{tabular}{|c|c|c|c|c|c|}
\hline
Model&TNR&TPR&FNR&FPR\\
\hline
RVAE&\textbf{0.784}&0.683&0.317&\textbf{0.215}\\
$with\_label$&0.681&0.918&0.082&0.319\\
\hline
$without\_label(0.05)$&0.649&0.885&0.115&0.351\\
$without\_label(0.1)$&0.716&0.899&0.101&0.284\\
$without\_label(0.5)$&0.632&\textbf{0.923}&\textbf{0.077}&0.368\\
\hline
\end{tabular}
\end{center}
\end{table*}

\noindent\textbf{Performance comparison with semi-supervised method.} We validate the proposed method by comparing with semi-supervised learning that uses RVAE method.
We find that our proposed method $with\_label$ outperforms RVAE, semi-supervised learning, as you can see in Table.~\ref{main_result}.
We highlight TPR among other metrics since the purpose of using the proposed model is to detect botnet.
TPR, which is also called detection rate, of $with\_label$ method is 0.918 while TPR of RVAE method is 0.683 in Table.~\ref{main_result}. Even we obtain higher detection rate with $without\_label$ method ($r_s$=0.1) (0.899).
This output indicates that the effectiveness of using transfer learning as we use the same RVAE model and set the same hyper-parameters for training. 
Moreover, even $without\_label$ method that does not use label information on the target domain shows higher performance than RVAE on TPR and FNR metrics. 
Overall, these results demonstrate that the proposed method detects suspicious botnet better on the target domain as using transferred knowledge that is obtained on the related domain (source domain) can provide useful information for the target domain lack of training data.

\noindent\textbf{Performance comparison with varying $r_s$ on without\_label method.} We perform experiments to find the optimal $r_s$ on $without\_label$ method. 
We change ratio of samples that are used again for the next batch training. 
In the case of dataset K, it has about 5\% botnet samples in total. We change $r_s$ from 3\% to 50\% for experiments. 
Setting the ratio as 50\% means that large portion of samples are used as normal samples for the next batch training. 
It leads to high TPR and low FNR as large portion of samples are re-used for training that makes reconstruction error distribution of normal samples biased to small reconstruction errors.
It can have many samples predicted as botnet.
Also, using high ratio causes low TNR and high FPR because of the same reason. 
On the other hand, when $r_s$ is 5\%, we obtain higher TNR and low FPR, comparatively. 
It is because we use the small portion of samples, which has little impact on distinguishing botnet from normal samples.
Therefore, setting the ratio appropriately is a key factor on using $without\_label$ method. 
We propose empirical methodology of setting ratio by doing several experiments changing $r_s$ in this paper.
For the purpose of practical use, the ratio should be determined by which measure is important.
 
\section{Conclusion}
This paper focuses on improving botnet detection performance with the consideration of two challenges: characterization of botnet traffic (showing sequential patterns) and the shortage of labeled datasets essentially required for constructing learning models. 
We first presented the RVAE anomaly detection method taking into account for the sequential and periodic nature for the network traffic flow data. 
The study is of significance to providing the applicable solution for the botnet detection system, especially in an online manner. 
To validate RVAE method, we apply the proposed method with the
CTU-13 dataset, a widely used dataset for botnet detection studies, and show that the proposed method can detect previously unseen
botnets more effectively by utilizing sequential patterns of the network traffic compared to the methodology without using sequential pattern as AUROC is 0.979 higher than 0.963 of MLP-VAE.
As the proposed method is validated on various scenarios of botnet operation, including the botnets that are not used for training, it can be concluded that the proposed method is robust in detecting previously unseen botnets. 

We also  presented a framework for transfer learning,
to overcome the challenge of the unavailability of labeled datasets.
Transfer learning can learn on the old data with labels and then apply the learning model on new data records as well as data records collected differently without labels.
The ability of working with unlabeled data is particularly useful for the network security applications because security issues such as botnets continue to evolve.
In our evaluation of the transfer learning framework,
we use CTU-13 dataset as the source domain for training and a fresh set of network monitoring data as the target domain.
Tests show that the proposed transfer learning method is able to detect
botnets better than a semi-supervised learning method trained on the target domain data. 
We observe that True Positive Rate (TPR) is 0.918 for transfer learning and 0.683 for directly using RVAE on the target domain data.
This indicates that the transfer learning could reliably identify anomalies. 

For future studies, we plan to study some improvements in the proposed method. 
From the perspective of RVAE structure, fuzzy logic can be adapted to improve the anomaly detector utilizing PDF.
It can provide more logical and systematic way of using PDFs for anomaly detection than comparing likelihoods from two distributions.
In addition, it is potential to improve performance of the anomaly detector if the method to cope with some cases of botnets having the small reconstruction errors from the normal cases is developed.
The common characteristics of the cases of botnets, which use a proprietary protocol, can be utilized to develop such a method. 
Moreover, the various VAE or RVAE architectures can be adapted to improve their anomaly detection performance. From the view of the transfer learning, while we propose an empirical manner of using transfer anomaly detection method without labels on a target domain, future research will be required to propose more systematic method beyond empirical ways to improve $without\_label$ method.
In addition, it is potential to improve performance of the anomaly detector in the FPR measure as it shows the weak performance relatively.

\section*{Acknowledgement}
This work was supported by the Office of Advanced Scientific Computing Research, Office of Science, of the U.S. Department of Energy under Contract No. DE-AC02-05CH11231, and also used resources of the National Energy Research Scientific Computing Center (NERSC). The authors also gratefully acknowledge Kimoon Jeong of National Supercomputing \& Networking, Korea Institute of Science and Technology Information (KISTI).

\bibliographystyle{ieeetr}
\bibliography{access.bib}

\end{document}